\documentclass{article}

\usepackage{microtype}
\usepackage{graphicx}
\usepackage{booktabs} 
\usepackage{algorithm}
\usepackage{cite}
\usepackage{amsmath,amssymb,amsfonts}

\usepackage{multirow}
\usepackage{acro}

\makeatletter
\def\BState{\State\hskip-\ALG@thistlm}
\makeatother

\usepackage{graphicx}
\usepackage{textcomp}
\usepackage{xcolor}
\usepackage{todonotes}
\usepackage{authblk}
\def\BibTeX{{\rm B\kern-.05em{\sc i\kern-.025em b}\kern-.08em
		T\kern-.1667em\lower.7ex\hbox{E}\kern-.125emX}}
\usepackage{cleveref}

\usepackage{pifont}
\usepackage{placeins}

\usepackage{float}
\restylefloat{table}

\newcommand{\cmark}{\ding{51}}%
\newcommand{\xmark}{\ding{55}}%

\usepackage{color}
\definecolor{gray}{rgb}{0.4,0.4,0.4}
\definecolor{darkblue}{rgb}{0.0,0.0,0.6}
\definecolor{cyan}{rgb}{0.0,0.6,0.6}

\newcommand{\cifar}{CIFAR10}
\newcommand{\imagenet}{ImageNet}
\newcommand{\wideresnetcif}{WideResNet28-10}
\newcommand{\wideresnetim}{WideResNet51-2}
\newcommand{\autoattack}{{\it AutoAttack}}
\newcommand{\mnist}{MNIST}
\newcommand{\etal}{\textit{et al.}}

\newcommand{\fscore}{$F1$}

\newcommand{\whitebox}{White-Box}
\newcommand{\blackbox}{Black-Box}

\newcommand{\apgdce}{APGD-CE}
\newcommand{\apgdt}{APGD-t}

\newcommand{\squaredef}{Squares}

\DeclareAcronym{cnn}{
  short=CNN,
  long=Convolutional Neural Networks,
}

\DeclareAcronym{at}{
  short=AT,
  long=Adversarial Training,
}

\DeclareAcronym{fnr}{
  short=FNR,
  long=False Negative Rate,
}

\DeclareAcronym{asr}{
  short=ASR,
  long=Adversarial Succes Rate,
}

\DeclareAcronym{asrd}{
  short=ASRD,
  long=Adversarial Success Rate under Detection,
}

\DeclareAcronym{bb}{
  short=BB,
  long=BlackBox,
}

\DeclareAcronym{wb}{
  short=WB,
  long=WhiteBox,
}

\DeclareAcronym{lid}{
  short=LID,
  long=Local Intrinsic Dimensionality,
}

\DeclareAcronym{mah}{
  short=M-D,
  long=Mahalanobis Distance,
}

\DeclareAcronym{sota}{
  short=SOTA,
  long=state-of-the-art,
}

\DeclareAcronym{dft}{
  short=DFT,
  long=Discrete Fourier Transformation,
}

\DeclareAcronym{fft}{
  short=FFT,
  long=Fast Fourier Transformation,
}

\DeclareAcronym{mfs}{
  short=MFS,
  long=magnitude Fourier spectrum,
}

\DeclareAcronym{pfs}{
  short=PFS,
  long=phase Fourier spectrum,
}

\DeclareAcronym{dnn}{
  short=DNN,
  long=Deep Neural Network,
}

\usepackage[accepted]{icml2021}

\icmltitlerunning{Detecting AutoAttack Perturbations in the Frequency Domain}

\begin{document}

\twocolumn[
\icmltitle{Detecting {\it \bf AutoAttack} Perturbations in the Frequency Domain} 




\begin{icmlauthorlist}
\icmlauthor{Peter Lorenz}{itwm,cvl}
\icmlauthor{Paula Harder}{itwm}
\icmlauthor{Dominik Straßel}{itwm}
\icmlauthor{Margret Keuper}{ma}
\icmlauthor{Janis Keuper}{itwm,imla}

\end{icmlauthorlist}

\icmlaffiliation{itwm}{Competence Center High Performance Computing, Fraunhofer ITWM, Kaiserslautern, Germany }
\icmlaffiliation{cvl}{Heidelberg University, Germany }
\icmlaffiliation{ma}{Data and Web Science Group, University of Mannheim, Germany }
\icmlaffiliation{imla}{Institute for Machine Learning and Analytics (IMLA), Offenburg University, Germany }

\icmlcorrespondingauthor{Peter Lorenz}{peter.lorenz@itwm.fhg.de}

\icmlkeywords{Adversarial Attack, Robustness, CNN}

\vskip 0.3in
]



\printAffiliationsAndNotice{\icmlEqualContribution} 

\begin{abstract}
    Recently, adversarial attacks on image classification networks by the {\it AutoAttack} \cite{autoattack} framework have drawn a lot of attention. While {\it AutoAttack } has shown a very high attack success rate, most defense approaches are focusing on network hardening and robustness enhancements, like adversarial training. This way, the currently best-reported method can withstand $\sim 66\%$ of adversarial examples on  \cifar.\\ 
    In this paper, we investigate the spatial and frequency domain properties of {\it AutoAttack} and propose an alternative defense. Instead of hardening a network, we detect adversarial attacks during inference, rejecting manipulated inputs. Based on a rather simple and fast analysis in the frequency domain, we introduce two different detection algorithms. First, a black box detector which only operates on the input images and achieves a detection accuracy of $100\%$ on the {\it AutoAttack} \cifar~ benchmark and $99.3\%$ on \imagenet, for $\epsilon=8/255$ in both cases. Second, a white-box detector using an analysis of \Acs{cnn} feature-maps, leading to a detection rate of also $100\%$ and $98.7\%$ on the same benchmarks.        
\end{abstract}

\section{Introduction}
The vulnerability of neural networks towards adversarial attacks is one of the key obstacles which are currently limiting the applicability of deep learning models for a wide range of practical use cases. While the latest attack methods, like \autoattack~\cite{Croce2020ReliableEO}, are achieving very high success rates perturbing input data on \acf{cnn} based image classifiers, the available defense methods appear to be ``always one step behind''. In general, adversarial defense strategies can be grouped into roughly two different approaches: First, the hardening of networks, which is mostly done via adversarial training, and second, the detection of adversarial samples during inference. In this work, we propose a simple feature-driven approach to detect adversarial examples, where features are extracted in the frequency domain, based on prior work by \cite{original}, which shows almost perfect detection results on \ac{sota} adversarial benchmarks.              

\begin{table}
\centering
\caption{Results of the proposed detectors on {\it AutoAttack (standard mode)} for different choices of the hyper-parameter $\varepsilon$ (default in most publications is $\varepsilon=8/255$) and test sets. \emph{ASR=Attack Success Rate}, \emph{ASRD=Attack Success Rate under Detection}. \whitebox~results on \imagenet~are obtained by a Logistic Regression classifier, Random Forests were used in all other cases. \Cref{tab:full_res} in the appendix is showing the full results for both classifiers. F1 and the False Negative Rate (FNR) are used to report the detection performance. See section \ref{sec:exp}  for details of the experimental setup.}
    \begin{tabular}{c@{\hspace{1mm}}c|@{\hspace{1mm}}|@{\hspace{1mm}}c@{\hspace{1mm}}c@{\hspace{1mm}}c|@{\hspace{1mm}}c@{\hspace{1mm}}c@{\hspace{1mm}}c} 
    
        \multicolumn{1}{c}{} & \multicolumn{1}{c|@{\hspace{1mm}}|@{\hspace{1mm}}}{} & \multicolumn{3}{c}{\blackbox} & \multicolumn{3}{c}{\whitebox} \\
        $\varepsilon$ &  ASR  &   F1  & FNR &    ASRD    &  F1   & FNR &  ASRD\\
        \hline 
         \multicolumn{8}{c}{\cifar~ on \wideresnetcif}  \\ \hline
        \textbf{8/255}   & \textbf{100}  &\textbf{98.2} & \textbf{00.0}  & \textbf{00.0}  & \textbf{99.0} &   \textbf{00.0}   &  \textbf{00.0} \\
        4/255   & 100  &        93.8  & 00.3  & 00.3  &         96.4  &   05.0   &  05.0 \\
        2/255   & 93.1 &        85.0  & 05.3  & 04.9  &         85.8  &   05.0   &  04.6 \\
        1/255   & 49.6 &        70.8  & 22.7  & 11.3  &         62.5  &   37.3   &  18.5 \\
        0.5/255 & 12.3 &        54.8  & 46.7  & 05.8  &         61.6  &   52.0   &  06.4 \\ 
        \hline
        \multicolumn{8}{c}{\imagenet~ on \wideresnetim } \\ \hline
        \textbf{8/255} & \textbf{100} & \textbf{87.1} & \textbf{00.7} &  \textbf{00.7} &   \textbf{96.7}  & \textbf{01.3} &  \textbf{01.3} \\
        4/255 & 100 &         77.4  & 08.7 &  08.7 &            94.1  & 02.7 &  02.7 \\
        2/255 & 100 &         60.2  & 28.0 &  28.0 &            82.3  & 16.7 &  16.7 \\
        1/255 & 99.9 &       53.4  &  43.7 &  43.6 &            67.9  & 30.7 &  30.6 \\
        0.5/255 & 96.9 &     54.4  &  42.0 &  40.7 &            59.0  & 41.0 &  38.1 \\
    \end{tabular}
    \label{tab:results}
\end{table}

\begin{figure*}[ht]
    \includegraphics[width=0.26\paperwidth]{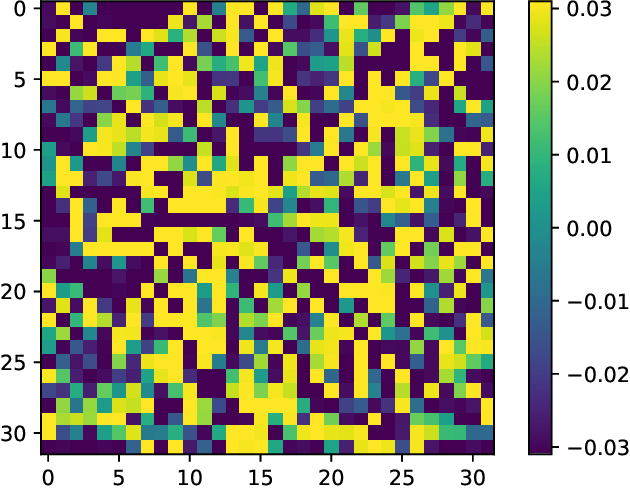}
    \includegraphics[width=0.26\paperwidth]{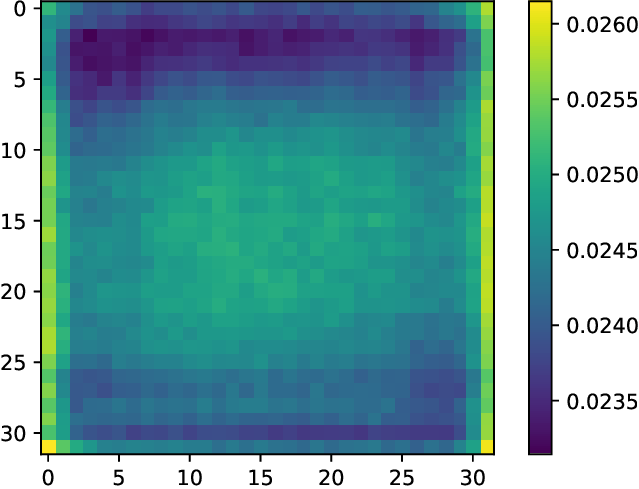}
    \includegraphics[width=0.26\paperwidth]{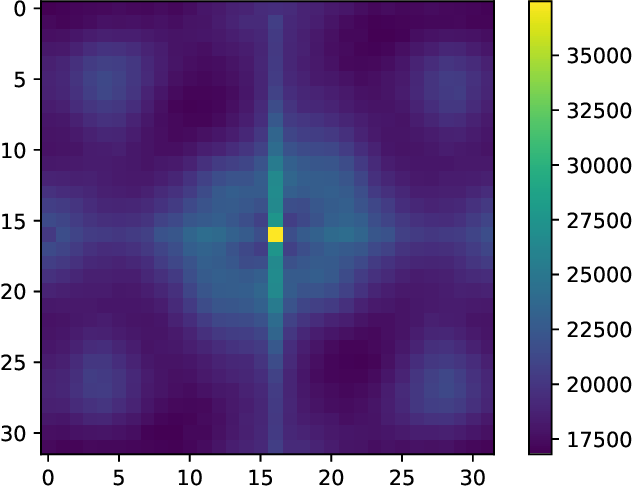}
    
    \includegraphics[width=0.26\paperwidth]{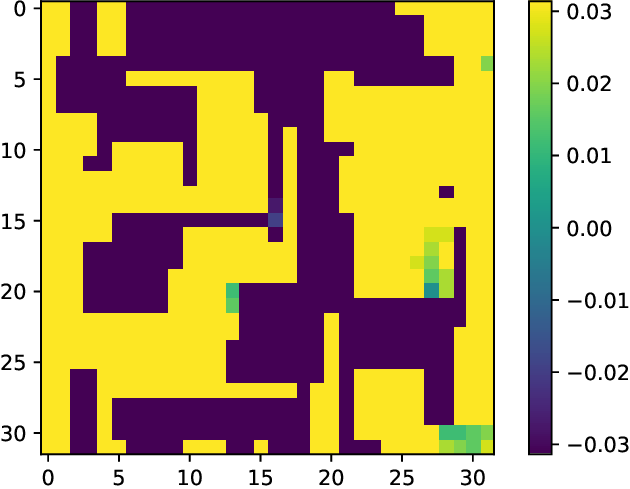}
    \includegraphics[width=0.26\paperwidth]{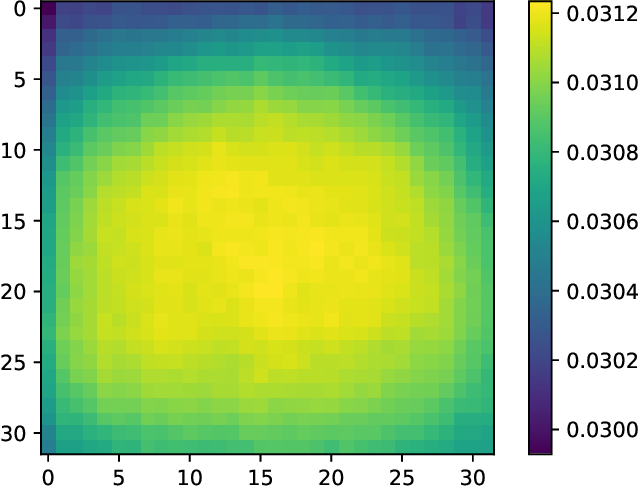}
    \includegraphics[width=0.26\paperwidth]{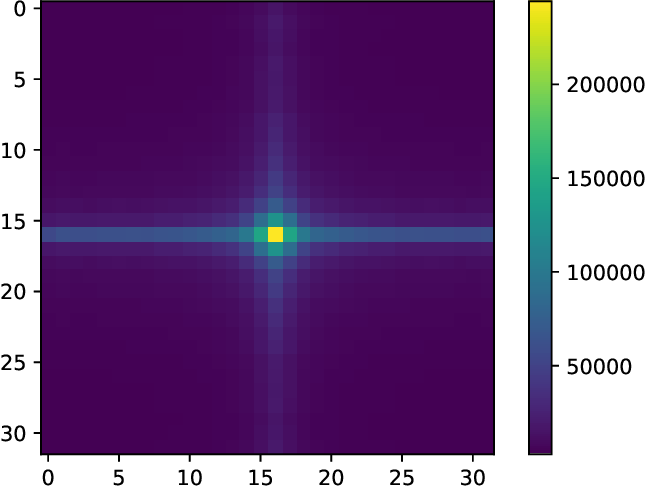}
    \caption{
    \label{fig:fft} Visualization of AutoAttack perturbations on a ResNet18 for \cifar. The top row: \apgdce~ $\ell_\infty$ attack, bottom row: Squares $\ell_\infty$ attack. Left column shows the spacial difference between a random test image from \cifar~and its perturbation. The center column depicts the mean of spacial differences over 1000 perturbed images. Right column: accumulated magnitudes of the spectral differences over the same 1000 images. While there are no obvious clues that can be obtained from the spacial domain, the frequency representation of perturbations show significant and systematic changes which can be exploited to detect attacks.    
    }
\end{figure*}

\subsection{Related Work}

\textbf{The AutoAttack Benchmark. }\label{rel_autoattack}In 2020, ~\cite{Croce2020RobustBench} launched a benchmark website\footnote{robustbench.github.io} with the goal to provide a standardized benchmark for adversarial robustness. Until then, single related libraries such as FoolBox \cite{foolbox}, Cleverhans \cite{papernot2018cleverhans} and AdverTorch \cite{2019advertorch} were already available but did not include all \ac{sota}~methods in one evaluation. \\
The currently dominating adversarial attack method is  \autoattack~ \cite{Croce2020ReliableEO} which is an ensemble of 4 attacks: two variations of the PGD \cite{pgd} attack with  cross-entropy loss (\apgdce) and difference of logits ratio loss (\apgdt), the targeted version of the FAB attack \cite{fabtattack}, and the black-box \squaredef~ attack \cite{squareattack}. 
The \autoattack~ benchmark provides several modes. The ``standard'' mode executes the 4 attack methods consecutively. Only if one attack fails, the failed samples are handed over to the next attack method.
The ``individual'' mode provides results for each attack on all input samples. 

\textbf{Adversarial Training.}\label{rel_ad_train}
\ac{at}  can be backtracked to \cite{fgsm}, in which models were hardened by generating adversarial examples and adding them into training data. An adversarial example is a subtly modified image causing a machine learning model to misclassify it. The achieved robustness by \ac{at} depends on the strength of the adversarial examples used. 
For example, training on Goodfellow's FGSM, which is fast and non-iterative, only provides robustness against non-iterative attacks, but e.g. not against 
PGD~\cite{pgd} attacks. Consequently, \cite{aaa} propose training on multi-step PGD adversaries, achieving \ac{sota} robustness levels against $\ell_\infty$ attacks on \mnist~ and \cifar~ datasets. Unfortunately, the high computational complexity of adversarial training  makes it impractical for large-scale problems such as \imagenet.

\begin{table*}[ht!]
    \resizebox{\textwidth}{!}{
    \begin{tabular}{lllllll}
   
    Rank &
      Method &
      \begin{tabular}[c]{@{}l@{}}Standard \\ Accuracy\end{tabular} &
      \begin{tabular}[c]{@{}l@{}}Robust\\ Accuracy\end{tabular} &
      \begin{tabular}[c]{@{}l@{}}Extra\\ data\end{tabular} &
      Architecture &
      Date  \\ \hline 
    1 &
      Fixing Data Augmentation to Improve Adversarial Robustness &
      92.23\% &
      66.56\% &
      \cmark &
      WideResNet-70-16 &
      Mar 2021 \\
    2 &
      Uncovering the Limits of Adversarial Training against Norm-Bounded Adversarial Examples &
      91.10\% &
      65.87\% &
      \cmark &
      WideResNet-70-16 &
      Oct 2020 \\
    3 &
       Fixing Data Augmentation to Improve Adversarial Robustness &
      88.50\% &
      64.58\% &
      \xmark &
      WideResNet-106-16 &
      Mar 2021 \\
    4 &
       Fixing Data Augmentation to Improve Adversarial Robustness &
      88.54\% &
      64.20\% &
      \xmark &
      WideResNet-70-16 &
      Mar 2021 \\
    5 &
      Uncovering the Limits of Adversarial Training against Norm-Bounded Adversarial Examples &
      89.48\% &
      62.76\% &
      \cmark &
      WideResNet-28-10 &
      Oct 2020 

    \end{tabular}
    }
    \caption{RobustBench: The top-5 entries of \cifar~ leaderboard for $\ell_\infty$ in June 2021.} 
    \label{tab:robustbench}
\end{table*}

\textbf{Adversarial Detection.}\label{rel_ad_det}
	Many recent works have focused on adversarial detection, trying to distinguish adversarial from natural images. The authors in \cite{pca_1} showed that adversarial examples have higher weights for larger principal components of the images' decomposition and use this finding to train a detector. Both \cite{pca_2} and \cite{pca_3} leverage Principal Component Analysis (PCA) as well.
	Based on the responses of the neural networks' final layer ~\cite{feinman} define two metrics, the kernel density estimation and the Bayesian neural network uncertainty to identify adversarial perturbations.
	\cite{steganalysis} proposed a method to detect adversarial examples by leveraging steganalysis and estimating the probability of modifications caused by adversarial attacks. \cite{grosse} used the statistical test of maximum mean discrepancy to detect adversarial samples. Using the correlation between images, based on influence functions and the k-nearest neighbors in the embedding space of the \ac{dnn}, \cite{nnif} proposed an adversarial detector.
	Besides the statistical analysis of the input images, adding a second neural network to decide whether an image is an adversarial example is another possibility. \cite{metzen} proposed a model that is trained on outputs of multiple intermediate layers. 
	Two strong and popular detectors are the \ac{lid}~\cite{lid} and the \ac{mah}~\cite{mah} detectors. Ma~\etal~used the \ac{lid} as a characteristic of adversarial subspaces and identified attacks using this measure. Lee~\etal~computed the empirical mean and covariance for each training sample and then calculated the \ac{mah} distance between a test sample and its nearest class-conditional Gaussians.

\textbf{Fourier Analysis of Adversarial Attacks.}
	\cite{fourier_uap} showed that \ac{cnn} are sensitive in the direction of Fourier basis functions, and proposed a Fourier-based attack method. Investigating trade-offs between Gaussian data augmentation and adversarial training \cite{fourier_robust} take a Fourier perspective and observe that adversarial examples are not only a high-frequency phenomenon. In \cite{benford} it is assumed that internal responses of \ac{dnn} follow the generalized Gaussian distribution, both for benign and adversarial examples, but with different parameters. They extract the feature maps at each layer in the classification network and calculate the Benford-Fourier coefficients for all of these representations. This approach is similar to our white-box detector, but as our experiments show, it is more than sufficient to use our simplified features built on a standard 2D \ac{dft}.

\section{Methods: \ac{dft} based Detection}
Our proposed detection method is based on the frequency-domain features originally introduced in \cite{original}, which we revise in the next subsections. In contrast to \cite{original}, we explicitly propose two types of detectors. First, a White-Box detector which has access to the feature maps of the target network, allowing it to observe the network response to input images, and second, a more general Black-Box detector which has no knowledge about the target network. We found that the Fourier power spectrum provides sufficient information to detect perturbations in both cases. Hence, we neglect the phase-based features which are also suggested in \cite{original}.

	\textbf{Fourier Analysis.} \label{sec:fft}
	The Fourier transformation decomposes a function into its constituent frequencies. A signal sampled at equidistant points is thereby known as discrete Fourier transform. The discrete Fourier transform of a signal with length $N$ can be computed efficiently with the \ac{fft} in $\mathcal{O}(N\log N)$ \cite{fft}. 
	For a discrete 2D signal, like color image channels or single CNN feature maps -- $X\in[0,1]^{N\times N}$ -- the 2D discrete Fourier transform is given as
	\begin{equation}\label{eq:eq1}
	    \mathcal{F}(X)(l,k) = \sum_{n,m=0}^N e^{-2\pi i \frac{lm+kn}{N}}X(m,n),
	\end{equation}
	for $l,k = 0,\ldots N-1$, with complex valued Fourier coefficients $\mathcal{F}(X)(l,k)$.
	In the following, we will only utilize the magnitudes of Fourier coefficients 
	\begin{equation}
	    |\mathcal{F}(X)(l,k)| = \sqrt{\text{Re}(\mathcal{F}(X)(l,k))^2 +\text{Im}(\mathcal{F}(X)(l,k))^2}
	    \label{eq:fftabs}
	\end{equation}
	and show that this information is sufficient to detect adversarial attacks with high accuracy.
	
	\subsection{\blackbox~Detection: Fourier Features of Input Images}
	
	\Cref{fig:fft} gives a brief visualization of the analysis of the changes in successfully perturbed images from {\it AutoAttack}: While different attacks show distinct but randomly located change patterns in the spatial domain (which makes them hard to detect), adversarial samples show strong, well localized signals in the frequency domain.\\
	Hence, we extract and concatenate the 2D power spectrum of each color channel (see \cref{eq:fftabs} and the right column of  \cref{fig:fft}) as feature representations of input images and use simple classifiers like Random Forests and Logistic Regression to learn to detect perturbed input images.
	
	\subsection{\whitebox~ Detection: Fourier Features of Feature-Maps}
	In the white-box case, we apply the same method as in the black-box approach, but extend the inputs to the feature map responses of the target network to test samples. Since this extension will drastically increase the feature space for larger target networks, we select only a subset of the available feature maps. Note that the optimal selection of feature maps heavily depends on the topology of the target network. See \cref{tab:features} for details on our selection for \cifar.

    \subsection{Measuring Adversarial Detection}
    The {\it AutoAttack} benchmark \cite{Croce2020ReliableEO}, like most of the literature regarding adversarial robustness, uses a ''Robust Accuracy'' measure to compare different methods (see \cref{tab:robustbench} for details). However, our approach does not fit this evaluation scheme, since we are aiming to reject adversarial test samples instead of hardening the networks. Therefore, we report two different indicators: 
    The {\it \ac{asr}} in \cref{eq:asr}  is calculated as
    \begin{equation}
        \text{ASR} = \frac{ \text{\#~perturbed~samples }}{ \text{\#~all~samples} } \label{eq:asr}
    \end{equation}
     the fraction of successfully perturbed test images and provides a baseline of \autoattack's ability to fool unprotected target networks. 

We measure the performance of our defense by the {\it \ac{asrd} } in \cref{eq:asrd}. Here, we compute the ratio of successful attacks under defense
\begin{equation}
    \text{ASRD} = \frac{ \text{\#~undetected~perturbations} } { \text{\#~all~samples} } = \text{FNR} \cdot \text{ASR,} \label{eq:asrd}
\end{equation}
where \Acs{fnr} is the false negative rate of the applied detection algorithm. \ac{asr} is only a scaling factor for the \ac{fnr}. The lower the \ac{asrd} rate, the more pertubated examples are defeated.

\section{Experiments} \label{sec:exp}


\textbf{CIFAR10.} We trained a \wideresnetcif~\cite{wideresidual} on the \cifar~training set to a test-accuracy of 94\% and applied \autoattack~on the test set. Then we extracted the spectral features and used a random subset of 1500 samples of this data for each attack method to train and evaluate our base classifiers (train:test split 80:20).\\ 
\Cref{tab:results} shows results using \autoattack~in ``standard'' mode for various $\varepsilon$ on $\ell_\infty$-perturbations. 
\Cref{tab:individ} shows the results for the ``individual'' mode for $\varepsilon=8/255$ and $\ell_\infty$-perturbations as well as a comparison to other detection methods. Here we used 1000 samples from each dataset, \cifar~and \imagenet, for our evaluation.

\begin{table}[h!]
    \small
    \caption{
    \fscore-score comparison of detection methods. The attacks from \autoattack individual mode are applied using the \cifar~test set on a \wideresnetcif~with  $\varepsilon=8/255$.  Logistic regression has been used as base classifier.
    }
    
    \scalebox{0.99}{
        \begin{tabular}{cc|cccc}
           & & \multicolumn{4}{c}{Attack} \\
        Dataset & Detector & apgd-ce & apgd-t & fab-t  & square \\ \hline
         \multicolumn{6}{l}{ \cifar~ on \wideresnetcif} \\ \hline
                                     & \blackbox    & 94 & 91 & {60}  & 72 \\
                                     & \whitebox    & \textbf{97} & \textbf{97} & 69 & \textbf{96} \\
                                     & LID          & 93 & 98 & 88 & 90 \\
                                     & \ac{mah}     & 97 & 92 & \textbf{91} & 94\\ \hline
       \multicolumn{6}{l}{ \imagenet~ on \wideresnetim}\\ \hline
                                     & \blackbox    & 82 &   77 & 60 & 78 \\
                                     & \whitebox    & \textbf{97} & 97 & 52 & \textbf{95} \\
                                     & LID          & 96 & 81  & 53 & 66 \\
                                     & \ac{mah}     & 96 & \textbf{99} & \textbf{98} & 94 \\ 
        \end{tabular} 
    }
    \label{tab:individ}
\end{table}

\begin{table}[h!]
\caption{ Comparison of individual features from different layers via \fscore-score. Individual AutoAttack. Base classifier is logistic regression. $\varepsilon=8/255$.}
\small
\center
\begin{tabular}{cc|cccc}
      \multicolumn{6}{c}{\cifar~ on \wideresnetcif}              \\ \hline
                  &        & \multicolumn{4}{c}{Attack} \\
    Layers & Dim. & \multicolumn{1}{c}{apgd-ce} & \multicolumn{1}{c}{apgd-t} & \multicolumn{1}{c}{fab-t} & \multicolumn{1}{c}{square} \\ \hline
   conv2 0 WB  & 32768  & 93             & 88                     & \textbf{58}                 & 75                           \\
   conv2 1 WB  & 327680 & 94             & 89                     & 55                          & 77                  \\
   conv2 2 WB  & 327680 & 93             & \textbf{93}            & 56                          & 75                   \\
   conv2 3 WB  & 327680 & 93             & 92                     & 57                       & \textbf{85}  \\
   conv3 0 WB  & 327680 & \textbf{96}    & 89                     & 56                           & 78                  \\
   conv3 1 WB  & 163840 & 77             & 64                     & 45                           & 67                   \\
   conv3 2 WB  & 163840 & 72             & 61                     & 47                           & 64                  \\
   conv3 3 WB  & 163840 & 75             & 65                     & 50                           & 69                  \\
   conv4 0 WB  & 163840 & 78             & 65                     & 45                           & 70                  \\
   conv4 1 WB  & 81920  & 69             & 54                     & 47                           & 58                  \\
   conv4 2 WB  & 81920  & 70             & 56                     & 48                           & 55                  \\
   conv4 3 WB  & 81920  & 68             & 56                     & 46                           & 61                  \\
   relu        & 81920  & 68             & 56                     & 50                           & 60                           
    \end{tabular}
    \label{tab:features}
\end{table}

\textbf{\imagenet. }
  For the benchmarks on \imagenet~ we used the pre-trained \wideresnetim~\cite{wideresidual} from the PyTorch library. As test set, we apply the official validation set from \imagenet. The accuracy of this pre-trained model is about~$78\%$. 
  As for \cifar, \autoattack~ also shows strong adversarial performance on \imagenet. \Cref{tab:results} shows very high  \ac{asr} even for low $\varepsilon$ values. Note that $\varepsilon<0.5$ would not represent realistic attack scenarios since saving the perturbed images would round the adversarial changes to the next of 256 available bins in commonly used 8-bit per channel image encodings. \Cref{tab:individ} shows the results for individual attacks.

\section{Discussion and Conclusion} 
In this paper, we are able to show a first proof of concept that simple frequency features can be used to detect current \ac{sota} attacks with a very high accuracy on the standard \cifar~benchmark and on a the more complex \imagenet~ dataset. Especially the black-box approach could provide a practical counter-measure for the defense of real-world applications.  
However, there are still many open questions: I) How well will the detectors generalize to other datasets, network architectures, and new attacks? II) Why is \autoattack~so successful for very small $\varepsilon$ in \imagenet? III) Can the detection be combined with \acl{at} methods like the ones shown in \cref{tab:robustbench}? In light of these open questions, we expect our approach can build a solid basis for future research.           




\bibliographystyle{icml2021}

\begin{thebibliography}{27}
\providecommand{\natexlab}[1]{#1}
\providecommand{\url}[1]{\texttt{#1}}
\expandafter\ifx\csname urlstyle\endcsname\relax
  \providecommand{\doi}[1]{doi: #1}\else
  \providecommand{\doi}{doi: \begingroup \urlstyle{rm}\Url}\fi

\bibitem[Andriushchenko et~al.(2020)Andriushchenko, Croce, Flammarion, and
  Hein]{squareattack}
Andriushchenko, M., Croce, F., Flammarion, N., and Hein, M.
\newblock Square attack: a query-efficient black-box adversarial attack via
  random search.
\newblock In \emph{ECCV}, 2020.

\bibitem[Bhagoji et~al.(2017)Bhagoji, Cullina, and Mittal]{pca_3}
Bhagoji, A., Cullina, D., and Mittal, P.
\newblock Dimensionality reduction as a defense against evasion attacks on
  machine learning classifiers.
\newblock \emph{arXiv}, abs/1704.02654, 2017.

\bibitem[Cohen et~al.(2020)Cohen, Sapiro, and Giryes]{nnif}
Cohen, G., Sapiro, G., and Giryes, R.
\newblock Detecting adversarial samples using influence functions and nearest
  neighbors.
\newblock \emph{CVPR}, pp.\  14441--14450, 2020.

\bibitem[Cooley \& Tukey(1965)Cooley and Tukey]{fft}
Cooley, J. and Tukey, J.
\newblock An algorithm for the machine calculation of complex fourier series.
\newblock \emph{Mathematics of Computation}, 19:\penalty0 297--301, 1965.

\bibitem[Croce \& Hein(2020{\natexlab{a}})Croce and Hein]{Croce2020ReliableEO}
Croce, F. and Hein, M.
\newblock Reliable evaluation of adversarial robustness with an ensemble of
  diverse parameter-free attacks.
\newblock In \emph{ICML}, 2020{\natexlab{a}}.

\bibitem[Croce \& Hein(2020{\natexlab{b}})Croce and Hein]{autoattack}
Croce, F. and Hein, M.
\newblock Reliable evaluation of adversarial robustness with an ensemble of
  diverse parameter-free attacks.
\newblock In \emph{ICML}, 2020{\natexlab{b}}.

\bibitem[Croce \& Hein(2020{\natexlab{c}})Croce and Hein]{fabtattack}
Croce, F. and Hein, M.
\newblock Minimally distorted adversarial examples with a fast adaptive
  boundary attack.
\newblock In \emph{ICML}, 2020{\natexlab{c}}.

\bibitem[Croce et~al.(2020)Croce, Andriushchenko, Sehwag, Debenedetti,
  Flammarion, Chiang, Mittal, and Hein]{Croce2020RobustBench}
Croce, F., Andriushchenko, M., Sehwag, V., Debenedetti, E., Flammarion, N.,
  Chiang, M., Mittal, P., and Hein, M.
\newblock Robustbench: a standardized adversarial robustness benchmark.
\newblock \emph{arXiv preprint arXiv:2010.09670}, 2020.

\bibitem[Ding et~al.(2019)Ding, Wang, and Jin]{2019advertorch}
Ding, G.~W., Wang, L., and Jin, X.
\newblock advertorch v0.1: An adversarial robustness toolbox based on pytorch.
\newblock \emph{arXiv}, abs/1902.07623, 2019.

\bibitem[Feinman et~al.(2017)Feinman, Curtin, Shintre, and Gardner]{feinman}
Feinman, R., Curtin, R.~R., Shintre, S., and Gardner, A.~B.
\newblock Detecting adversarial samples from artifacts.
\newblock \emph{arXiv}, abs/1703.00410, 2017.

\bibitem[Goodfellow et~al.(2015)Goodfellow, Shlens, and Szegedy]{fgsm}
Goodfellow, I., Shlens, J., and Szegedy, C.
\newblock Explaining and harnessing adversarial examples.
\newblock \emph{CoRR}, abs/1412.6572, 2015.

\bibitem[Grosse et~al.(2017)Grosse, Manoharan, Papernot, Backes, and
  Mcdaniel]{grosse}
Grosse, K., Manoharan, P., Papernot, N., Backes, M., and Mcdaniel, P.
\newblock On the (statistical) detection of adversarial examples.
\newblock \emph{arXiv}, abs/1702.06280, 2017.

\bibitem[Harder et~al.(2021)Harder, Pfreundt, Keuper, and Keuper]{original}
Harder, P., Pfreundt, F.-J., Keuper, M., and Keuper, J.
\newblock Spectraldefense: Detecting adversarial attacks on cnns in the fourier
  domain.
\newblock \emph{arXiv preprint arXiv:2103.03000}, 2021.

\bibitem[Hendrycks \& Gimpel(2017)Hendrycks and Gimpel]{pca_1}
Hendrycks, D. and Gimpel, K.
\newblock Early methods for detecting adversarial images.
\newblock \emph{arXiv: Learning}, 2017.

\bibitem[Lee et~al.(2018)Lee, Lee, Lee, and Shin]{mah}
Lee, K., Lee, K., Lee, H., and Shin, J.
\newblock A simple unified framework for detecting out-of-distribution samples
  and adversarial attacks.
\newblock In \emph{NeurIPS}, 2018.

\bibitem[Li \& Li(2017)Li and Li]{pca_2}
Li, X. and Li, F.
\newblock Adversarial examples detection in deep networks with convolutional
  filter statistics.
\newblock In \emph{ICCV}, pp.\  5775--5783, 2017.
\newblock \doi{10.1109/ICCV.2017.615}.

\bibitem[Liu et~al.(2019)Liu, Zhang, Zhang, Hou, Liu, Zha, and
  Yu]{steganalysis}
Liu, J., Zhang, W., Zhang, Y., Hou, D., Liu, Y., Zha, H., and Yu, N.
\newblock Detection based defense against adversarial examples from the
  steganalysis point of view.
\newblock \emph{CVPR}, pp.\  4820--4829, 2019.

\bibitem[Ma et~al.(2020)Ma, Wu, Xu, Fan, Zhang, Zhang, and Li]{benford}
Ma, C., Wu, B., Xu, S., Fan, Y., Zhang, Y., Zhang, X., and Li, Z.
\newblock Effective and robust detection of adversarial examples via
  benford-fourier coefficients.
\newblock \emph{arXiv}, abs/2005.05552, 2020.

\bibitem[Ma et~al.(2018)Ma, Li, Wang, Erfani, Wijewickrema, Houle, Schoenebeck,
  Song, and Bailey]{lid}
Ma, X., Li, B., Wang, Y., Erfani, S., Wijewickrema, S., Houle, M., Schoenebeck,
  G., Song, D., and Bailey, J.
\newblock Characterizing adversarial subspaces using local intrinsic
  dimensionality.
\newblock \emph{arXiv}, abs/1801.02613, 2018.

\bibitem[Madry et~al.(2018)Madry, Makelov, Schmidt, Tsipras, and Vladu]{pgd}
Madry, A., Makelov, A., Schmidt, L., Tsipras, D., and Vladu, A.
\newblock Towards deep learning models resistant to adversarial attacks.
\newblock \emph{arXiv}, abs/1706.06083, 2018.

\bibitem[Metzen et~al.(2017)Metzen, Genewein, Fischer, and Bischoff]{metzen}
Metzen, J.~H., Genewein, T., Fischer, V., and Bischoff, B.
\newblock On detecting adversarial perturbations.
\newblock \emph{arXiv}, abs/1702.04267, 2017.

\bibitem[Papernot et~al.(2016)Papernot, Faghri, Carlini, Goodfellow, Feinman,
  Kurakin, Xie, Sharma, Brown, Roy, Matyasko, Behzadan, Hambardzumyan, Zhang,
  Juang, Li, Sheatsley, Garg, Uesato, Gierke, Dong, Berthelot, Hendricks,
  Rauber, Long, and Mcdaniel]{papernot2018cleverhans}
Papernot, N., Faghri, F., Carlini, N., Goodfellow, I., Feinman, R., Kurakin,
  A., Xie, C., Sharma, Y., Brown, T., Roy, A., Matyasko, A., Behzadan, V.,
  Hambardzumyan, K., Zhang, Z., Juang, Y.-L., Li, Z., Sheatsley, R., Garg, A.,
  Uesato, J., Gierke, W., Dong, Y., Berthelot, D., Hendricks, P., Rauber, J.,
  Long, R., and Mcdaniel, P.
\newblock Technical report on the cleverhans v2.1.0 adversarial examples
  library.
\newblock \emph{arXiv: Learning}, 2016.

\bibitem[Rauber et~al.(2018)Rauber, Brendel, and Bethge]{foolbox}
Rauber, J., Brendel, W., and Bethge, M.
\newblock Foolbox: A python toolbox to benchmark the robustness of machine
  learning models.
\newblock \emph{arXiv}, abs/1707.04131, 2018.

\bibitem[Tram{\`e}r et~al.(2020)Tram{\`e}r, Carlini, Brendel, and Madry]{aaa}
Tram{\`e}r, F., Carlini, N., Brendel, W., and Madry, A.
\newblock On adaptive attacks to adversarial example defenses.
\newblock \emph{arXiv}, abs/2002.08347, 2020.

\bibitem[Tsuzuku \& Sato(2019)Tsuzuku and Sato]{fourier_uap}
Tsuzuku, Y. and Sato, I.
\newblock On the structural sensitivity of deep convolutional networks to the
  directions of fourier basis functions.
\newblock \emph{CVPR}, pp.\  51--60, 2019.

\bibitem[Yin et~al.(2019)Yin, Lopes, Shlens, Cubuk, and Gilmer]{fourier_robust}
Yin, D., Lopes, R.~G., Shlens, J., Cubuk, E.~D., and Gilmer, J.
\newblock A fourier perspective on model robustness in computer vision.
\newblock \emph{arXiv}, abs/1906.08988, 2019.

\bibitem[Zagoruyko \& Komodakis(2017)Zagoruyko and Komodakis]{wideresidual}
Zagoruyko, S. and Komodakis, N.
\newblock Wide residual networks.
\newblock In \emph{BMVC}, 2017.

\end{thebibliography}

\pagebreak
\onecolumn
\section*{Appendix}

\begin{table*}[h!]

\caption{ Extension of \cref{tab:results}, showing full results for Logistic Regression (LR) and Random Forest (RF). RF outperforms LR in almost all cases except the White-Box approach on ImageNet. Due to the large feature-space, RF appear to be overfitting in this case.   \label{tab:full_res}}
    \center
    \begin{tabular}{cccccccccccccc}
    
     & \multicolumn{1}{c||}{} & \multicolumn{6}{c|}{\blackbox} & \multicolumn{6}{c}{\whitebox} \\
    \multirow{2}{*}{$\varepsilon$} & \multicolumn{1}{c||}{\multirow{2}{*}{ASR}} & \multicolumn{2}{c|}{F1} & \multicolumn{2}{c|}{FNR} & \multicolumn{2}{c|}{ASRD} & \multicolumn{2}{c|}{F1} & \multicolumn{2}{c|}{FNR} & \multicolumn{2}{c}{ASRD} \\ \cline{3-14} 
     & \multicolumn{1}{c||}{} & LR & \multicolumn{1}{|c|}{RF} & LR & \multicolumn{1}{|c|}{RF} & LR & \multicolumn{1}{|c|}{RF} & LR & \multicolumn{1}{|c|}{RF} & LR & \multicolumn{1}{|c|}{RF} & LR &  \multicolumn{1}{|c}{RF} \\ \hline
    \multicolumn{14}{c}{\cifar~on \wideresnetcif} \\ \hline
    8/255 & \multicolumn{1}{c||}{100} & 98.0 & \textbf{98.2} & 00.0 & \textbf{00.0} & 00.0 & \multicolumn{1}{c|}{\textbf{00.0}} & 97.2 & \textbf{99.0} & 02.7 & \textbf{00.0} & \multicolumn{1}{l}{02.7} & \textbf{00.0} \\
    4/255 & \multicolumn{1}{c||}{100} & \textbf{94.8} & 93.8 & 15.0 & \textbf{00.3} & 15.0 & \multicolumn{1}{c|}{\textbf{00.3}} & 83.7 & \textbf{96.4} & \multicolumn{1}{l}{18.0} & \textbf{05.0} & \multicolumn{1}{l}{18.0} & \textbf{05.0} \\
    2/255 & \multicolumn{1}{c||}{93.1} & \textbf{86.3} & 85.0 & 28.7 & \textbf{05.3} & 26.7 & \multicolumn{1}{c|}{\textbf{04.9}} & 62.9 & \textbf{85.8} & 35.0 & \textbf{05.0} & \multicolumn{1}{l}{32.6} & \textbf{04.6} \\
    1/255 & \multicolumn{1}{c||}{49.6} & \textbf{74.0} & 70.8 & 41.3 & \textbf{22.7} & 20.5 & \multicolumn{1}{c|}{\textbf{11.3}} & 54.8 & \textbf{62.5} & 46.0 & \textbf{37.3} & 22.8 & \textbf{18.5} \\
    0.5/255 & \multicolumn{1}{c||}{12.3} & 54.4 & \textbf{54.8} & 48.3 & \textbf{46.7} & 06.0 & \multicolumn{1}{c|}{\textbf{05.8}} & 51.7 & \textbf{61.6} & \textbf{50.0} & 52.0 & \textbf{06.0} & 06.4 \\ \hline
    \multicolumn{14}{c}{\imagenet~on \wideresnetim} \\ \hline
    8/255 & \multicolumn{1}{c||}{100} & 81.9 & \textbf{87.1} & 16.3 & \textbf{00.7} & 16.3 & \multicolumn{1}{c|}{\textbf{00.7}} & \textbf{96.7} & 90.4 & \textbf{01.3} & 03.0 & \textbf{01.3} & 03.0 \\
    4/255 & \multicolumn{1}{c||}{100} & 65.0 & \textbf{77.4} & 36.0 & \textbf{08.7} & 36.0 & \multicolumn{1}{c|}{\textbf{08.7}} & \textbf{94.1} & 82.3 & \textbf{02.7} & 07.7 & \textbf{02.7} & 07.7 \\
    2/255 & \multicolumn{1}{c||}{100} & 55.9 & \textbf{60.2} & 44.3 & \textbf{28.0} & 44.3 & \multicolumn{1}{c|}{\textbf{28.0}} & \textbf{82.3} & 73.5 & \textbf{16.7} & 20.7 & \textbf{16.7} & 20.7 \\
    1/255 & \multicolumn{1}{c||}{99.9} & 50.8 & \textbf{53.4} & 50.7 & \textbf{43.7} & 50.6 & \multicolumn{1}{c|}{\textbf{43.6}} & \textbf{67.9} & 59.1 & \textbf{30.7} & 40.7 & \textbf{30.6} & 40.6 \\
    0.5/255 & \multicolumn{1}{c||}{96.9} & \textbf{54.7} & 54.4 & \textbf{40.7} & 42.0 & \textbf{39.4} & \multicolumn{1}{c|}{40.7} & \textbf{59.0} & 50.5 & \textbf{41.0} & 50.0 & \textbf{38.1} & 48.5
    \end{tabular}
\end{table*}

\end{document}